%% file: root.tex
\documentclass[letterpaper, 10 pt, conference]{ieeeconf}  % Comment this line out
\IEEEoverridecommandlockouts                              % This command is only
\usepackage{graphics} % for pdf, bitmapped graphics files
\usepackage{epsfig} % for postscript graphics files
\usepackage{float}
\usepackage{mathptmx} % assumes new font selection scheme installed
\usepackage{times} % assumes new font selection scheme installed
\usepackage{amsmath} % assumes amsmath package installed
\usepackage{amssymb}  % assumes amsmath package installed

\usepackage{url}
\usepackage[misc]{ifsym}
\usepackage{tikz}
\usepackage{tikz-qtree}
\usepackage{multirow}
\usetikzlibrary{arrows}
\usetikzlibrary{arrows.meta}
\usetikzlibrary{decorations.markings}
\usetikzlibrary{decorations.pathreplacing}
\usetikzlibrary{arrows,shapes,automata,petri,positioning,calc}
\usepackage{adjustbox}
\usepackage{bbding}
\usepackage{xcolor}
\usepackage[english]{babel}
\usepackage{bm}
\usepackage{verbatim}
\usepackage{relsize}

\usepackage{subfiles} % load last in the preamble

\newcommand{\family}[1]{\mathcal{#1}}

\DeclareMathOperator*{\argmax}{argmax}

\DeclareMathOperator*{\eqdef}{\stackrel{\text{\tiny def}}{=}}

\newtheorem{theorem}{Theorem}

\newtheorem{proposition}[theorem]{Proposition}
\newtheorem{definition}[theorem]{Definition}

\newtheorem{claim}[theorem]{Claim}
\newtheorem{lemma}[theorem]{Lemma}

\title{\LARGE \bf
Lower Bounds for Policy Iteration on Multi-action MDPs
}

\author{Kumar Ashutosh$^{\dagger}$, Sarthak Consul$^{\dagger}$, Bhishma Dedhia$^{\dagger}$, Parthasarathi Khirwadkar$^{\dagger}$, Sahil Shah$^{\dagger}$,\\  and Shivaram Kalyanakrishnan\Envelope% <-this % stops a space
\thanks{$^{\dagger}$ indicates equal contribution. \Envelope \hspace{0.2em} indicates corresponding author. All authors are affiliated to the Indian Institute of Technology Bombay, Mumbai, India. E-mail: {\tt\small \{kumar.ashutosh@, sarthakconsul@, bhishma@, parthasarathi.k@, sahilshah@cse., shivaram@cse.\}iitb.ac.in}. }%
}

\begin{document}
\maketitle
\thispagestyle{empty}
\pagestyle{empty}

%%%%%%%%%%%%%%%%%%%%%%%%%%%%%%%%%%%%%%%%%%%%%%%%%%%%%%%%%%%%%%%%%%%%%%%%%%%%%%%%
\begin{abstract}
Policy Iteration (PI) is a classical family of algorithms to compute an optimal policy for any given Markov Decision Problem (MDP). The basic idea in PI is to begin with some initial policy and to repeatedly update the policy to one from an improving set, until an optimal policy is reached. Different variants of PI result from the (switching) rule used for improvement. An important theoretical question is how many iterations a specified PI variant will take to terminate as a function of the number of states $n$ and the number of actions $k$ in the input MDP. While there has been considerable progress towards upper-bounding this number, there are fewer results on lower bounds. In particular, existing lower bounds primarily focus on the special case of $k = 2$ actions. We devise lower bounds for $k \geq 3$. Our main result is that a particular variant of PI can take $\Omega(k^{n/2})$ iterations to terminate. We also generalise existing constructions on $2$-action MDPs to scale lower bounds by a factor of $k$ for some common deterministic variants of PI, and by $\log(k)$ for corresponding randomised variants.
%Policy Iteration (PI) algorithm is an iterative method to obtain the optimal policy for a given Markov Decision Problem (MDP). The lower bound for a particular PI variant is the maximum number of iterations required to converge to the optimal policy of a given MDP starting from an initial policy. It provides us with a measure of the tightness of the upper bound, and gives us an idea about the worst-case time complexity of the algorithm. In this paper, we consider the problem of finding a PI variant, called Adversarial PI, whose lower bound is close to the trivial upper bound of $k^n$ for an $n$-state $k$-action MDP. We theoretically prove that the Adversarial PI takes $O(k^{0.5n})$ when applied to the proposed  $n$-state $k$-action MDP. Further, we extend previous results for well-known PI variants on 2-action MDPs to arbitrary number of actions. A theoretical justification is provided along with each proposed family of MDPs.
\end{abstract}

%%%%%%%%%%%%%%%%%%%%%%%%%%%%%%%%%%%%%%%%%%%%%%%%%%%%%%%%%%%%%%%%%%%%%%%%%%%%%%%%
\section{Introduction}\label{sec:intro}
Markov Decision Problems (MDPs) \cite{Bellman1957}\cite{Puterman1994} are a popular abstraction of sequential decision making tasks in stochastic environments. An MDP is a tuple $\langle S, A, T, R, \gamma \rangle $, where $S$ is a set of states and $A$ is a set of actions. $T: S \times A \times S \rightarrow [0,1]$ is a function such that $T(s, a, s^{\prime})$ is the probability of reaching state $s^{\prime} \in S$ from state $s \in S$ by taking action $a \in A$. The reward function $R: S \times A \rightarrow \mathbb{R}$, assigns a bounded reward $R(s, a)$ when the agent takes action $a \in A$ from state $s \in S$. 

An MDP serves as an environment, which describes the consequences of an agent's actions. The agent itself has control only over its own behaviour, encapsulated as a \textit{policy} $\pi:S \rightarrow A$ (by this definition, policies are Markovian, stationary, and deterministic---sufficient for our purposes). If $r_{0}, r_{1}, r_{2}, \dots$ denotes the sequence of rewards obtained by an agent that follows policy $\pi$, starting at state $s \in S$, then its expected long-term reward
\begin{equation}
\label{eqn:value-def}
    V^{\pi}(s) \eqdef \mathbb{E}_{\pi, s}[r_{0} + \gamma r_{1} + \gamma^{2} r_{2} + \dots]
\end{equation}
is denoted the \textit{value} of $s$ under $\pi$; $V^{\pi}: S \to \mathbb{R}$ is the \textit{value function} of $\pi$.
In \eqref{eqn:value-def}, $\gamma \in [0, 1]$ is a discount factor. In general, $\gamma$ is set to be less than $1$ so that the \textit{infinite discounted reward} is well-defined. However, we may set $\gamma = 1$, thereby taking value to be the \textit{total reward}, when trajectories in the input MDP are guaranteed to reach a terminal state. In this paper, we adopt the total reward formulation, but our results can all be extended to the infinite discounted setting.

Every MDP is guaranteed to have an \textit{optimal} policy $\pi^{\star}:S \to A$ whose value at each state is at least as large as any other policy's~\cite{Bellman1957}. Hence, given an MDP, a natural objective is to compute an optimal policy for it. There are many approaches to this planning problem, among them Value Iteration and Linear Programming~\cite{Littman+DK:1995}. In this paper, we consider a third popular approach: Policy Iteration (PI).

PI~\cite{howard:dp} is based on the Policy Improvement Theorem, which facilitates a relatively straightforward computation of a set of locally-improving policies ${\textbf{IP}}(\pi)$ for any given policy $\pi$. ${\textbf{IP}}(\pi)$ is represented implicitly through ``improvable states'' for $\pi$, as well as ``improving actions'' for such states. If $\pi$ is optimal, ${\textbf{IP}}(\pi)$ is guaranteed to be empty; if not, every policy $\pi^{\prime} \in {\textbf{IP}}(\pi)$ strictly dominates $\pi$ in terms of state values. Indeed every such policy $\pi^{\prime} \in {\textbf{IP}}(\pi)$ is obtained by switching the actions taken by $\pi$ in some improvable states to corresponding improving actions.

Given an arbitrary initial policy $\pi_{0}$, a PI algorithm generates a sequence of policies $\pi_{0}, \pi_{1}, \dots, \pi_{T}$ wherein $\pi_{t + 1} \in {\textbf{IP}}(\pi_{t})$ for $t = 0, 1, \dots, T - 1$, and $\pi_{T}$ is an optimal policy. Even for the same MDP and starting policy $\pi_{0}$, different PI variants could select improving policies in different ways, thereby yielding different sequences. In this paper, our aim is to \textit{lower-bound} the length of these sequences. We restrict our attention to finite MDPs, assuming that $S$ shall comprise $n$ non-terminal states and a constant number of terminal states. We take $A = \{0, 1, \dots, k - 1\}$; thus $|A| = k$. With this setup, observe that policies can be viewed as $n$-length $k$-ary strings. 

Since PI increases some state value in each iteration, it cannot visit the same policy more than once. Hence, $k^{n}$, which is the total number of policies, serves as a trivial upper bound on the iterations taken by every PI variant. Howard's PI \cite{howard:dp}, a classical variant, has been shown to incur no more than $O(k^{n}/n)$ iterations \cite{mansour_singh}. Among upper bounds that are solely in terms of $n$ and $k$, the tightest are
$O(k^{0.7019n})$ iterations for deterministic PI variants~\cite{Taraviya2019}, and
$O\left((2 + \ln(k - 1))^{n}\right)$ expected iterations for randomised variants~\cite{Kalyanakrishnan+MG:2016}. Even tighter upper bounds (still exponential in $n$) have been shown for $k = 2$~\cite{Taraviya2019}. Interestingly, the only \textit{lower} bounds that have been shown for PI are either for the special case of $k = 2$~\cite{Melekopoglou1994}\cite{HZ2010} or when $k$ is related to  $n$~\cite{Fearnley2010}\cite{Hollanders+DJ:2012}. We contribute lower bounds for arbitrary $n \geq 2$, $k \geq 2$.

Every PI variant must choose which improvable states to switch. Notably, this is all that PI needs to do on $2$-action MDPs, since selecting an improvable state fixes the improving action. The main technical difference that arises on $k$-action MDPs, $k \geq 3$, is that there can be multiple improving actions associated with an improvable state, and PI must additionally choose among them. We consider both deterministic and randomised strategies for action selection. Our main contribution is a novel MDP construction that yields a trajectory of length $\Omega(k^{n/2})$ for a particular deterministic variant of PI. From a theoretical perspective, it is significant that the base of the exponent is an increasing (in fact polynomial) function of $k$. We also generalise existing constructions for $2$-action MDPs, scaling lower bounds by a factor of $k$ for some deterministic PI variants, and by $\log(k)$ for some randomised variants. We present our constructions in sections \ref{sec:api}--\ref{sec:spi}, after first formalising PI in Section~\ref{sec:pi} and discussing existing lower bounds in Section~\ref{sec:existingLB}. We present conclusions and discuss future directions in Section~\ref{sec:conclusions}.

\section{Policy Iteration}\label{sec:pi}

In this section, we describe Policy Iteration (PI), borrowing notation from previous work~\cite{Taraviya2019}\cite{Kalyanakrishnan+MG:2016}. Note that for any given policies $\pi$ and $\pi^{\prime}$, the relation $\pi \succeq \pi^{\prime}$ means that for all $s \in S$, $V^{\pi}(s) \geq V^{\pi^{\prime}}(s).$ If $\pi \succeq \pi^{\prime}$, and for some $s \in S$, $V^{\pi}(s) > V^{\pi^{\prime}}(s)$, then we also have $\pi \succ \pi^{\prime}$.\\

\noindent\textbf{Policy evaluation.} Each iteration of PI considers some policy $\pi$, and begins by computing its value function $V^{\pi}$. From the definition in \eqref{eqn:value-def}, it is seen that $V^{\pi}$ satisfies a set of linear equations (called Bellman's Equations): for $s \in S$,
\vspace{-0.05cm}
\begin{equation*} \label{eq:bellman}
   V^{\pi}(s) = R(s, \pi(s)) + \gamma \sum_{s^{\prime} \in S} T(s, \pi(s), s^{\prime}) V^{\pi}(s^{\prime}).
\end{equation*}
The ``action value function'' of $\pi$, $Q^\pi: S \times A \rightarrow \mathbb{R}$, is defined as follows: for $s \in S, a \in A$, $Q^{\pi}(s, a)$ is the expected long-term reward the agent receives if it takes action $a$ from state $s$ for the first time-step, and then follows policy $\pi$. Thus,
\vspace{-0.05cm}
\begin{equation*} \label{eq:Q_defn}
Q^{\pi}(s, a) = R(s, a) + \gamma \sum_{s^{\prime} \in S} T(s, a, s^{\prime}) V^{\pi}(s^{\prime}).
\end{equation*}
\noindent\textbf{Policy improvement.} Define $\text{\bf IS}(\pi)$ to be the set of states $s$ on which $\pi$ is \textit{not} greedy with respect to its own action-value function: that is,
\vspace{-0.05cm}
\begin{equation*}               \text{\bf IS}(\pi) \eqdef \left\{s \in S: Q^{\pi}(s, \pi(s)) < \max_{a    \in A} Q^{\pi}(s, a)\right\}.     \end{equation*}
For each state $s\in \text{\bf IS}(\pi)$, the set of improving actions $\text{\bf IA}(\pi, s)$ is defined as:
\vspace{-0.05cm}
\begin{equation*}
\text{\bf IA}(\pi, s)\eqdef \left\{a \in A: Q^{\pi}(s, a) > Q^{\pi}(s, \pi(s))\right\}.
\end{equation*}
% \doubt{Need to define IP}
If $\text{\bf IS}(\pi)$ is not empty, let $\pi^{\prime}$ be a policy that takes some action from $\text{\bf IA}(\pi, s)$ for one or more states $s \in \text{\bf IS}(\pi)$, and takes the same action as $\pi$ in the remaining states. In other words, $\pi^{\prime}$ satisfies
\vspace{-0.05cm}
\begin{align}
&\exists s \in S: \pi^{\prime}(s) \in \text{\bf IA}(\pi, s), \text{ and}\nonumber \\           &\forall s \in S: (\pi^{\prime}(s) = \pi(s)) \vee (\pi^{\prime}(s) \in \text{\bf IA}(\pi, s)). \label{eq:policyimprovement}
\end{align}
Denote the set of all $\pi^{\prime}$ satisfying \eqref{eq:policyimprovement} as the set $\text{\bf IP}(\pi)$:
\vspace{-0.05cm}
\begin{equation}
    {\bf IP}(\pi) \eqdef \{\pi^{\prime} \in \Pi: \pi^{\prime} \text{ satisfies \eqref{eq:policyimprovement}}\}. \nonumber
\end{equation}
The Policy Improvement Theorem shows that every policy $\pi^{\prime} \in \text{\bf IP}(\pi)$ improves upon (or \textit{dominates}) $\pi$ as follows.
\begin{theorem}[Policy improvement]
\label{thm:policyimprovement}
For every $\pi: S \to A$:\\(1) if $\text{\bf IS}(\pi) \neq \emptyset$, then for all $\pi^{\prime} \in \text{\bf IP}(\pi)$, $\pi^{\prime} \succ \pi$;\\(2) if $\text{\bf IS}(\pi) = \emptyset$, then for all $\pi^{\prime}: S \to A$, $\pi \succeq \pi^{\prime}$.
\end{theorem}
\vspace{0.2cm}
The proof of this well-known theorem is available from many sources~\cite{Kalyanakrishnan+MG:2016}\cite{Bertsekas+Tsitsiklis:1996}.
\vspace{0.1cm}

\noindent\textbf{Switching rules.} For a given policy $\pi$, it is immediate that $\text{\bf IS}(\pi)$ and $\text{\bf IA}(\pi, \cdot)$---which implicitly represent $\text{\bf IP}(\pi)$---can be computed using $\text{poly}(n, k)$ arithmetic operations. The overall running-time of the algorithm may therefore be obtained by multiplying this per-iteration complexity with the total number of iterations taken to terminate. In turn, the number of iterations is determined by the rule used to pick $\pi^{\prime} \in \text{\bf IP}(\pi)$ as the policy following $\pi$.

Recall that $\pi^{\prime}$ is obtained by modifying $\pi$: by selecting one or more states from $s \in \text{\bf IS}(\pi)$, and switching to some action from $\text{\bf IA}(\pi, s)$ for such states $s$. The most common variant of PI, called Howard's PI or Greedy PI \cite{howard:dp}, switches \textit{every} state $s \in \text{\bf IS}(\pi)$. By contrast, under the Random PI variant~\cite{mansour_singh}, a non-empty subset of $\text{\bf IS}(\pi)$ is selected uniformly at random, and the states within this subset are switched. Under Simple PI~\cite{Melekopoglou1994}, which is yet another variant, only a single improvable state is switched. Assuming a fixed indexing of states for the entire run of the algorithm, in each iteration the improvable state with the largest index is switched.

In $2$-action MDPs, it suffices to specify which states to switch, since an improvable state will have exactly one improving action. On the other hand, if there are $k \geq 3$ actions, one might encounter improvable states with multiple improving actions, requiring yet another decision to be made.

\begin{itemize}
    \item A common strategy for action-selection is to pick an action that maximises the $Q$-value: that is, setting $\pi^{\prime}(s) \gets \argmax_{a \in A} Q^{\pi}(s, a)$ for a selected improvable state $s \in \text{\bf IS}(\pi)$.
    In this paper, we are unable to furnish meaningful lower bounds for this ``max-Q'' strategy. We make headway with two other natural approaches.
    \item Our first, ``index-based'' action-selection strategy assumes a fixed indexing of actions for the entire run of the algorithm, and always switches to the improving action with the smallest index. Since we have assumed $A = \{0, 1, \dots, k - 1\}$, we set $\pi^{\prime}(s) \gets \min (\text{\bf IA}(\pi, s)).$
    \item Our second, ``random'' strategy  sets $\pi^{\prime}(s)$ to an action picked uniformly at random from $\text{\bf IA}(\pi, s)$.
\end{itemize}
We couple these action-selection strategies with several state-selection strategies and then lower-bound the number of iterations taken by the resulting PI variants. Before presenting our contributions, we review existing lower bounds for PI.

\section{Existing Lower Bounds}\label{sec:existingLB}

For $n$-state, $2$-action MDPs, Melekopoglou and Condon~\cite{Melekopoglou1994} show that Simple PI can take $\Omega(2^{n})$ iterations to terminate.
%They construct an MDP with $n$ non-terminal states, with each having a deterministic and a stochastic action. For convenient handling of the stochastic action, which can have as many as $\Theta(n)$ possible next states, their construction associates a dummy ``state'' (in which no decision is to be made) with each non-terminal state. Every action decreases takes the agent closer to termination, upon which either a negative or zero reward is received. Under the total reward formulation, the authors show a trajectory for Simple PI that visits all $2^{n}$ policies. In Section X, w
In Section~\ref{sec:spi}, we generalise both their construction and their proof to $k \geq 2$, obtaining lower bounds of $\Omega(k \cdot 2^{n})$ and 
$\Omega(\log(k) \cdot 2^{n})$ when Simple PI is applied with index-based and random action selection, respectively.

%We give full details of the generalisation in .

The tightest lower bounds known for Howard's PI~\cite{HZ2010} and Random PI~\cite{Taraviya2019} on $n$-state, $2$-action MDPs are only $\Omega(n)$. Hansen and Zwick \cite{HZ2010} construct a deterministic MDP on which, under the ``average reward'' criterion~\cite{Mahadevan:1996}, Howard's PI can take as many as $2n - O(1)$ iterations. %We focus on a specific portion of the resulting trajectory and generalise it to $k \geq 2$ under the total reward setting.
We show linear dependence on $n$ using a simpler construction, and obtain linear and logarithmic scaling in $k$ for index-based and random action selection, respectively (see Section~\ref{sec:hpi}).
%\shivaram{Does the Hansen and Zwick lower bound also hold for all PI variants: that is, does it also have only one improvable state at a time? Or is it only your adaptation that has this property?} \reply{Answer: No. This property is true only in our adaptation. In Hansen Zwick's construction, there are more than one improvable states in some of the policies.}
Interestingly, our construction also implies a lower bound of $\Omega(kn)$ (or $\Omega(\log(k) \cdot n)$) iterations for index-based (respectively, random) action selection regardless of the state-selection strategy used. %Hence, we match the linear lower bound shown recently for Random PI [].

Indeed
a trajectory of exponential length ($\Omega(2^{n/7})$) has been shown for Howard's PI both under the total reward~\cite{Fearnley2010} and infinite discounted reward~\cite{Hollanders+DJ:2012} settings. However, the MDPs used in these constructions do not have a constant number of actions per state---rather, this number is itself  $\theta(n)$. Yet another exponential lower bound (of $\Omega(2^{n/2})$ iterations) has been shown for Howard's PI on a class of objects called Acyclic Unique Sink Orientations (AUSOs), which may be derived from $n$-state, $2$-action MDPs~\cite{Schurr+Szabo:2005}. The proof does not imply the same bound for MDPs~\cite{Taraviya2019}. 

The bounds mentioned above, and also the ones we provide, only depend on $n$ and $k$. While there are upper bounds for PI in terms of parameters such as the discount factor, we are not aware of any such lower bounds.

\section{A Trajectory of Length $\Omega(k^{n/2})$}\label{sec:api}

In this section, we propose a novel family of $n$-state, $k$-action MDPs on which a particular variant of PI can take $\Omega(k^{n/2})$ iterations to terminate. This lower bound becomes the tightest shown yet for the PI family. In subsequent sections, we generalise lower bounds for 
specific, commonly-used variants of PI to 
$k \geq 2$, but the resulting bounds are only linear or logarithmic in $k$.

\subsection{Construction of Family $F(m, k)$}\label{sec:api_construction}

We construct a family of MDPs with $n = 2m$ non-terminal states, $m \geq 1$, a single terminal state, and $k$-actions, as shown in Fig.~\ref{fig:adversarial_family2}. The idea behind the construction is to implement a $k$-ary ``counter'' on a set of non-terminal states $s_{1}, s_{2}, \dots, s_{m}$, ensuring that all $k^{m}$ sub-policies on these states are visited. To this end, we employ a ``partner'' state $s^{\prime}_{i}$ for each such state $s_{i}$, $i \in \{1, 2, \dots, m\}$. Recall that $A = \{0, 1, \dots, k - 1\}$.

As shown in Fig.~\ref{fig:adversarial_family2}, all transitions in $F(m, k)$ are deterministic. Moreover, each state $s_i$ in the counter and its partner $s^{\prime}_i$ have identical next states and rewards for each action. From state $s_{1}$ all actions $j \in A$ lead to the terminal state $s_{T}$. From state $s_{i}$, $i \in \{2, 3, \dots, m\}$, action $0$ alone leads to $s^{\prime}_{i - 1}$, while actions $j \in A \setminus \{0\}$ all lead to $s_{i - 1}$. For $i \in \{1, 2, \dots, m\}, j \in A$, the associated reward is $R(s_{i}, j) = jk^{m - i}$. Observe that there can be at most $m$ transitions before termination; no discounting is used in the calculation of values.

\subfile{API_construct_inv}

\subsection{Policies}\label{sec:api_pol}
We find it convenient to denote policies for $F(m, k)$ in the form $x \cdot y$, where $x, y \in A^{m}$. In this notation, the sequence $x = x_{1}x_{2} \dots x_{m}$ lists the actions taken from states $s_{1}, s_{2}, \dots, s_{m}$, respectively, and $y = y_{1}y_{2} \dots y_{m}$ does the same for states $s^{\prime}_{1}, s^{\prime}_{2}, \dots, s^{\prime}_{m}$, respectively. For every $x \in A^{m}$ and $r \in \{0, 1, 2, \dots, m\}$, let $\texttt{pre}(x:r)$ denote the prefix sequence $x_{1} x_{2} \dots x_{r}$.
%, and $\texttt{suf}(x:r)$ denote the suffix sequence $x_{r + 1} x_{r + 2} \dots x_{m}$. These (possibly empty) sequences may be viewed as sub-policies acting on either the counter states or the partner states, as specified.
This (possibly empty) sequence may be viewed as a sub-policy on the counter states or the partner states.

Our proof relies on associating numbers with policies. For every sequence $x = x_{1}x_{2} \dots x_{r}$, where $r \geq 1$ and $x_{u} \in A$
for $u \in \{1, 2, \dots, r\}$, let $[x]$ denote the natural number represented in base $k$ by $x$: that is, $[x] \eqdef \sum_{u = 1}^{r} x_{u} k^{r - u}.$ Let $N$ denote the set of numbers $\{0, 1, \dots, k^{m} - 1\}$. It is immediately clear that $A^{m}$, which is the set of $m$-length $k$-ary sequences, is in $1$-to-$1$ correspondence with $N$, each $x \in A^{m}$ associated with $[x] \in N$.

Of especial interest to us is policies of the form $x \cdot x$ for $x \in A^{m}$: we refer to such policies as \textit{balanced} policies. Since every counter state $s_{i}$ and its partner $s^{\prime}_{i}$, $i \in \{1, 2, \dots, m\}$, have the same outgoing transitions and rewards in $F(m, k)$, it follows that $V^{x\cdot x}(s_{i}) = V^{x\cdot x}(s^{\prime}_{i})$. Moreover, since all transitions either terminate or move to states with lower indices, these values only depend on $\texttt{pre}(x:i)$. Incorporating the corresponding rewards, we observe:
\begin{equation}
\label{eq:statevalue}
V^{x\cdot x}(s_{i}) = V^{x\cdot x}(s^{\prime}_{i}) = \sum_{u = 1}^{i} x_{i} k^{m - u} = k^{m - i}[\texttt{pre}(x:i)],
\end{equation}
and in particular, $V^{x\cdot x}(s_{m}) = V^{x\cdot x}(s^{\prime}_{m}) = [x]$. The format in \eqref{eq:statevalue} is convenient to establish a key property of $F(m, k)$.
\vspace{0.2cm}
\begin{proposition}[Comparability of balanced policies]
\label{prop:totalorder}
For $x, y \in A^{m}$, if $[y] > [x]$, then $y \cdot y \succ x \cdot x$.
\end{proposition}
\begin{proof}
``$[y] > [x]$'' is equivalently stated as: ``there exists $r \in \{0, 1, 2, \dots, m - 1\}$ such that for $u \in \{0, 1, \dots, r\}$, $[\texttt{pre}(y:u)] = [\texttt{pre}(x:u)]$ and for $u \in \{r + 1, r + 2, \dots, m\}$, $[\texttt{pre}(y:u)] > [\texttt{pre}(x:u)]$. From \eqref{eq:statevalue}, it follows that for $i \in \{1, 2, \dots, r\}$, $V^{y\cdot y}(s_{i}) = V^{y\cdot y}(s^{\prime}_{i}) = V^{x\cdot x}(s_{i}) = V^{x\cdot x}(s^{\prime}_{i}),$  and for $i \in \{r + 1, r + 2, \dots, m\}$, $V^{y\cdot y}(s_{i}) = V^{y\cdot y}(s^{\prime}_{i}) > V^{x\cdot x}(s_{i}) = V^{x\cdot x}(s^{\prime}_{i}),$
in turn implying that $y \cdot y \succ x \cdot x$.
\end{proof}
\vspace{0.2cm}

The proposition is seen to induce a total order on policies of the form $x \cdot x$ via their value functions. The maximal element, $k^{m} \cdot k^{m}$, is also the sole optimal policy for $F(m, k)$. Our proof will construct a trajectory for PI that visits each balanced policy; notice that there are $k^{m} = k^{n/2}$ in total.

At this point, one might wonder why we need the partner states in $F(m, k)$ at all. Consider an MDP
$F^{\prime}(m, k)$ that results from removing partner states from $F(m, k)$ and redirecting their incoming transitions to corresponding counter states. On $F^{\prime}(m, k)$, there would be a total order on the \textit{entire} set of $k^{m}$ polices, suggesting the possibility of an even tighter---in fact maximally tight---lower bound. However, crucially, it does not appear possible to get any \textit{PI variant} to visit all $k^{m}$ policies in $F^{\prime}(m, k)$. Although PI guarantees a dominating policy after each step, it is not necessary that every policy $\pi^{\prime}$ that dominates $\pi$ is reachable from $\pi$ using PI. With partner states, indeed we are able to show a chain of length $k^{m}$ for PI, but consequently $m$ is only half the number of states.

\subsection{A Long Trajectory for PI}\label{sec:api_traj}

We now present the main structural property of $F(m, k)$: that there is a sequence of policy improvements from every non-optimal balanced policy to its successor.
\vspace{0.2cm}
\begin{lemma}[Segments of Long PI Trajectory]
\label{lem:segments}
Consider $x, y \in A^{m}$ such that $[y] = [x] + 1$. There is a sequence of policies $\pi_{1}, \pi_{2}, \dots, \pi_{t + 1}$, $t \geq 2$, for $F(m, k)$ such that $\pi_{1} = x \cdot x$; 
 $\pi_{t + 1} = y \cdot y$; and for $i \in \{1, 2, \dots, t\}$, $\pi_{i + 1} \in \text{\bf IP}(\pi_{i})$.
\end{lemma}
\begin{proof}
We furnish a proof by showing a chain of policy improvements from $x \cdot x$ to $x \cdot y$, and another chain from $x \cdot y$ to $y \cdot y$. For the proof, we find it useful to denote as $I(z)$, for $z \in A^{m} \setminus \{(k - 1)^{m}\}$, the largest index of $z$ whose value is not $k - 1$. Also, we write the concatenation of sequences $z_{1}$ and $z_{2}$ as $z_{1}~\#~z_{2}$. With this notation, $[y] = [x] + 1$ implies 
\begin{align*}
x &= \texttt{pre}(x: I(x) - 1)~\#~x_{I(x)}~\#~(k - 1)^{m - I(x)}, \text{ and}\nonumber\\
y &= \texttt{pre}(x:I(x) - 1)~\#~x_{I(x)} + 1~\#~0^{m - I(x)}.\nonumber
\end{align*}
We show that PI can lead from $x \cdot x$ to $x \cdot y$ by switching, in sequence, the states $s^{\prime}_{I(x)}, s^{\prime}_{I(x) + 1}, \dots, s^{\prime}_{m}$; thereafter, switching $s_{m},
s_{m - 1}, \dots, s_{I(x)}$ in sequence leads from $x \cdot y$ to $y \cdot y$. Concretely, for $r \in \{1, 2, \dots, m - I(x) + 1\}$, define 
\begin{align*}
p_{r} &\eqdef \texttt{pre}(x:I(x) - 1)~\#~x_{I(x)} + 1~\#~0^{r - 1}~\#~(k - 1)^{m - I(x) - r + 1}\text{;}\nonumber\\
q_{r} &\eqdef \texttt{pre}(x:I(x) - 1)~\#~x_{I(x)}~\#~(k - 1)^{m - I(x) - r + 1}~\#~0^{r - 1}.
\end{align*}
We establish that the following sequences of policy improvements can be performed.
\begin{eqnarray}
x \cdot x \to x \cdot p_{1} \to x \cdot p_{2} \to \dots  \to x \cdot  p_{m - I(x) + 1} = x \cdot y \text{;}\nonumber \\ x \cdot y = q_1 \cdot y \to q_2 \cdot y  \to \dots  \to q_{m - I(x) + 1} \cdot y \to y \cdot y.\nonumber
\end{eqnarray}
For the first chain, observe that $x \cdot x$ and $x \cdot p_{1}$ differ only in one action: on state $s^{\prime}_{I(x)}$, $x \cdot x$ takes action $x_{I(x)}$ and $x \cdot p_{1}$ takes $x_{I(x)} + 1$. Using the structure of $F(m, k)$ and \eqref{eq:statevalue}, we get
\begin{align*}
Q^{x \cdot x}(s^{\prime}_{I(x)}, x_{I(x)} + 1)
&= (x_{I(x)} + 1)k^{m - I(x)} + V^{x \cdot x}(s_{I(x) - 1})\\
&> x_{I(x)} k^{m - I(x)} + V^{x \cdot x}(s_{I(x) - 1})\\
&= V^{x \cdot x}(s^{\prime}_{I(x)}),\nonumber
\end{align*}
with the convention that $s_{0} = s^{\prime}_{0} = s_{T}$. Now, for $r \in \{1, 2, \dots, m - I(x)\}$, policies $x \cdot p_{r}$ and $x \cdot p_{r + 1}$ take actions $k - 1$ and $0$ at state $s^{\prime}_{I(x) + r}$, respectively, but on other states are alike. Substituting values calculated using the structure of $F(m, k)$, we get
\begin{comment}
First, since $x$ and $p_r$ differ only at positions $I(x), I(x) + 1, \dots , I(x) + r - 1$, 
\begin{align}
    [\texttt{pre}(p_r:&I(x) + r - 1)] - [\texttt{pre}(x:I(x) + r - 1)] \nonumber\\ 
    &= \left[\texttt{pre}(x:I(x) - 1)~\#~x_{I(x)} + 1~\#~0^{r - 1}\right] \nonumber\\  
    &- \left[\texttt{pre}(x:I(x) - 1)~\#~x_{I(x)}~\#~(k - 1)^{r - 1}\right]\nonumber\\ 
    &= [x_{I(x)} + 1~\#~0^{r - 1}] - [~x_{I(x)}~\#~(k - 1)^{r - 1}]\nonumber\\
     &   = 1 \label{eq:diff_in_pre}
\end{align}
\begin{align}
        \therefore V^{x \cdot p_r}(s^{\prime}_{I(x) + r - 1})& - V^{x \cdot p_r}(s_{I(x) + r - 1}) \nonumber\\
        &= (k^{m - I(x) - r + 1})[\texttt{pre}(p_r:I(x) + r - 1)]\nonumber\\
        &- [\texttt{pre}(x:I(x) + r - 1)]\nonumber\\
        &= (k^{m - I(x) - r + 1})~\textnormal{from (\ref{eq:diff_in_pre})} \label{eq:diff_in_v}
\end{align}
\end{comment}
\begin{align*}
Q^{x \cdot p_r}(s^{\prime}_{I(x) + r}, 0) &= 0 + V^{x \cdot p_r}(s^{\prime}_{I(x) + r - 1}) \\
        & = V^{x \cdot p_r}(s_{I(x) + r - 1}) + k^{m - I(x) - r + 1}  \\
        &> V^{x \cdot p_r}(s_{I(x) + r - 1}) + (k-1)\cdot k^{m - I(x) - r}\\
        & = V^{x \cdot p_r}(s_{I(x) + r}).
\end{align*}
Intuitively, action $0$ is improving because the decrease in immediate reward on switching from action $k - 1$ to $0$ at state $s^{\prime}_{I(x) + r}$ is offset by the gain from moving to state $s^{\prime}_{I(x) + r - 1}$ instead of $s_{I(x) + r - 1}$. Recall that counter states follow $x$, while partner states follow $p_{r + 1}$, with a higher-index action at $I(x)$.

For the second chain we first show that for $r \in \{1,2, \dots , m - I(x)\}$, $q_{r + 1} \cdot y$ is improvable over $q_r \cdot y$. Note that the two policies take actions $0$ and $k -1$ at state $s_{m - r + 1}$ respectively, but are alike at all other states. Substituting values based on $F(m,k)$, we get
\begin{align*}
        Q^{q_r \cdot y}(s_{m - r + 1}, 0) 
        &= 0 + V^{q_r \cdot y}(s^{\prime}_{m - r}) \\
        &= k^r + V^{q_r \cdot y}(s_{m - r})  \\
        & > (k-1) \cdot k^{r-1} + V^{q_r \cdot y}(s_{m - r})\\
        & = V^{q_r \cdot y}(s_{m - r + 1}).
\end{align*}
Lastly, we need to show that the policy $y \cdot y$ improves over $q_{m - I(x) + 1} \cdot y$. Since the policies differ only at $s_{I(x)}$, showing
\begin{align*}
Q^{q_{m - I(x) + 1} \cdot x}&(s_{I(x)}, x_{I(x)} + 1)\\
&= (x_{I(x)} + 1)k^{m - I(x)} + V^{q_{m - I(x) + 1} \cdot x}(s_{I(x) - 1})\\
&> x_{I(x)} k^{m - I(x)} + V^{q_{m - I(x) + 1} \cdot x}(s_{I(x) - 1})\\
&= V^{q_{m - I(x) + 1} \cdot x}(s_{I(x)})
\end{align*}
concludes the proof.
\end{proof}
% \vspace{0.2cm}
\par In short, we have demonstrated that a sequence of policy improvements, each switching only a single state, can take us from $x \cdot x$ to $y \cdot y$. We denote the variant of PI that facilitates such a trajectory \textit{Peculiar PI}. For illustration, Appendix~\ref{appendix:f33trajectory} shows the sequence of policies visited by Peculiar PI on $F(3, 3)$. While it might appear that going from $x \cdot x$ to $y \cdot y$ requires keeping an intermediate sequence of policies in memory, indeed Peculiar PI can be implemented concisely as a memoryless variant, as shown in Appendix \ref{appendix:memory_less_algo}. From Lemma~\ref{lem:segments}, it is clear that if initialised with policy $0^{m}\cdot0^{m}$, this variant will visit all
$k^{m}$ balanced policies. 
\vspace{0.1cm}
\begin{theorem}[$\Omega(k^{n/2})$ Lower Bound for Peculiar PI]
On $F(m, k)$, if initialised with policy $0^{m} \cdot 0^{m}$, Peculiar PI takes $\Omega(k^{m})$ iterations.
\end{theorem}
\vspace{0.1cm}
\begin{comment}
The exact number of iterations can be computed by considering 2 different cases in in the trajectory:
\begin{enumerate}
    \item When $I(x) = m$, it takes $2$ iterations to increment the counter. There are $(k-1)k^{m-1}$ such $x \in A^m$. This results in the first term in (\ref{eqn:api_exact_steps}). 
    \item When $I(x) < m$, it takes $2(m-I(x)+1)$ iterations to increment the counter. There are $(k-1)k^{I(x)-1}$ such $x \in A^m$. This results in the second term in (\ref{eqn:api_exact_steps}). 
\end{enumerate}
which yields the following expression for the exact number of iterations taken:
\begin{align}\label{eqn:api_exact_steps}
    &2(k-1)k^{m-1} + 2(k-1)\sum_{i=1}^{m-1} (m-i+1)k^{i-1}\\ \nonumber
    &= 2(k-1)k^{m-1} + 2(k-1)\sum_{i=2}^m ik^{m-i}\\ \nonumber
    &= 2\left[\frac{k}{k - 1}\left(k^m  - 1\right) - m\right]
\end{align}

%% In the first case, incrementing the counter $x$ only involves changing the action at $s_m$, which takes 2 iterations. This occurs $(k-1)k^{m-1}$ times in the trajectory leading to the first term in (\ref{eqn:api_exact_steps}).
% In the second case, incrementing the counter causes an overflow, requiring in multiple steps to appropriately change the action of multiple states.
\end{comment}

Although this lower bound---and those from the next two sections---are shown using the total reward setting, they continue to hold with  discounting (see  Appendix~\ref{appendix:total_to_discounted}).

\section{Generic Lower Bounds}\label{sec:hpi}

In this section, we give $k$-dependent lower bounds for \textit{every} PI variant that uses index-based or random action selection; that is, the state-selection strategy can be arbitrary.

%The proposed construction of MDP is inspired from \cite{HZ2010}, where the problem of finding Minimum Mean-Cost Cycle in directed weighted graph is presented. 
% Our proposed construction is such that only state is improvable in an iteration of PI.
%With our specific modification to the construction, 
%In this section we provide a construction that has only one improvable state per policy. Hence, this lower bound applies to every index-based PI variant no matter how it picks the states to improve.
%\shivaram{Say some more to explain the difference between Hansen's construction and yours.}\doubt{Written in the next subsection ``CONSTRUCTION"}
%Here, we only consider PI variants that pick actions deterministically or uniformly at random. We give a bound in each case. Crucially, our bound also applies to Howard, which is used widely in practice, but has so far only been analysed for $k = 2$ in \cite{HZ2010}.

\subsection{Construction}\label{sec:hpi_construction}

Fig.~\ref{fig:gpi_construction} shows our family of MDPs $G(n, k)$ with non-terminal states $s_{1}, s_{2}, \dots, s_{n}$. Rewards are only given on reaching terminal states, of which there are $n + 1$.\footnote{If $\rho(s^{\prime})$ is the reward given on reaching $s^{\prime} \in S$ in addition to reward $R(s, a)$ given for taking action $a$ from state $s$, we can use $R^{\prime}(s, a) = R(s, a) + \sum_{s^{\prime} \in S} T(s, a, s^{\prime}) \rho(s^{\prime})$ as an equivalent reward function that complies with our definition in Section~\ref{sec:intro}. We use this idea here and in Section~\ref{sec:spi}.} From each state $s_{i}$, $i \in \{1, 2, \dots, n\}$, action $0$ deterministically terminates with a reward of $-2^{i}$. On the other hand, action $k - 1$ moves deterministically from $s_{i}$ to $s_{i + 1}$ for $i \in \{1, 2, \dots, n - 1\}$, and moves $s_{n}$ into a terminal state with no reward.

As before, let us denote policies as $n$-length, $k$-ary strings. Observe that for $i \in \{1, 2, \dots, n\}$, the policy $0^{i}(k - 1)^{n - i}$ has exactly one improvable state: $s_i$. If the only actions were $0$ and $k - 1$, any PI variant initialised with $0^{n}$ would be forced to visit all $n$ of these policies. To get PI to also take more actions from $\{1, 2, \dots, k - 2\}$, we implement \textit{stochastic} transitions for each of these actions. In particular, action $j \in A \setminus \{0, k - 1\}$ behaves like $0$ with probability $p_{j}$, and like $k - 1$ with probability $1 - p_{j}$, where $p_{j} = \frac{1}{2} + \frac{k - j}{2k}$. The intuition behind this construction is that (1) so long as state $s_{i + 1}$, $i \in \{1, 2, \dots, n - 1\}$ follows any action other than $k - 1$, action $0$ is the most rewarding at $s_{i}$; (2) once $s_{i + 1}$ switches to $k - 1$, actions in $A \setminus \{0\}$ become profitable at $s_{i}$. Concretely, we obtain the following structure within the set of policies.
\vspace{0.2cm}

\begin{lemma}\label{lemma:hpi_is_ia}
    For $i \in \{1, 2, \dots, n\}$, $j \in \{0, 1, \dots, k - 2\}$ let $\pi_{ij} = 0^{i - 1}j(k - 1)^{n - i}.$ Then 
    \begin{align*}
    \text{\bf IS}(\pi_{ij}) &= \{i\} \text{, and} \nonumber\\
    \text{\bf IA}(\pi_{ij}, i) &= \{j + 1, j + 2, \dots, k - 1\}.
    \end{align*}
\end{lemma}
It is straightforward to construct the proof by writing out and comparing $Q$-values in $G(n, k)$, as shown in Appendix~\ref{sec:hpilemmaproof}.

From Lemma~\ref{lemma:hpi_is_ia}, it follows directly that if initialised with the policy $0^{n}$, index-based action selection will go through $0^{n - 1}1, 0^{n - 1}2, \dots, 0^{n - 1}(k - 1)$; thereafter
$0^{n - 2}1(k - 1), 0^{n - 2}2(k - 1), \dots, 0^{n - 2}(k - 1)^{2}$, and so on until the optimal policy $(k - 1)^{n}$ is evaluated after $n(k - 1) + 1$ iterations.

In case random action selection is used, it remains that the policies $0^{n - 1}(k -1), 0^{n - 2}(k -1)^{2}, \dots, (k - 1)^{n}$ will be visited, but the number of policies visited in between any successive pair of these will be random, since improving actions are picked uniformly at random. For $i \in \{1, 2, \dots, n\}$, $j \in \{0, 1, \dots, k - 2\}$, let $t_{ij}$ denote the expected number of iterations needed to go from $\pi_{ij} = 0^{i - 1}j(k - 1)^{n - i}$ to $0^{i - 1}(k - 1)^{n - i + 1}$. Clearly $t_{ij}$ is independent of $i$, and may be written as $t_{j}$. We have $t_{k - 2} = 1$ and for $j \in \{0, 1, \dots, k - 3\}, t_{j} = 1 + \frac{1}{k - j - 1} \sum_{j^{\prime} = j + 1}^{k - 2} t_{j^{\prime}}$. Solving this recurrence yields $t_{0} = \theta(\log(k))$; in other words, there are $\theta(\log(k))$ expected iterations corresponding to each improvable state.
\vspace{0.1cm}
\begin{theorem}[Generic Lower Bounds]
On $G(n, k)$, if initialised with policy $0^{n}$, every PI variant doing index-based action selection takes $\Omega(kn)$ iterations, and every PI variant doing  random action selection takes $\Omega(\log(k) \cdot n)$ iterations.
\end{theorem}
\vspace{0.1cm}
This result is significant for Howard's PI and Random PI, whose current lower bounds are $\theta(n)$ even for $k = 2$.

%Old proof starts
% Since $R(s_q, k-1) > R(s_q, k-2) > \dots R(s_q, 1)$ and next state in all these actions is the same, it is enough to compare $Q(s_q, 0)$ and $Q(s_q, k-1)$. $Q(s_q, k-1) = 0$ and $Q(s_q, 0) = V_{\pi}(s_{q-1}) + R(s_q, 0) < 0$ for all $q>i$. Consequently, q \notin $\text{ \bf IS}(\pi)$ for all $q>i$.

% In a similar argument, $Q(s_i, t) > Q(s_i, j)$ for all $t>j$. $Q(s_i, 0) = R(s_i, 0) + V_{\pi}(s_{i-1}) = \sum_{q=1}^{i} R(s_q, 0) = 0.5(k-1)^{(i-1)} - (k-1)^i < R(s_i, j)$ which implies $0^{th}$ action is not improvable. Hence, i \in  $\text{ \bf IS}(\pi)$ and $\text{ \bf IA}(\pi, i) = \{j + 1, j + 2, \dots, k - 1\}$

% Similarly, $Q(s_q, 0) = \sum_{q=1}^{p} R(s_q, 0) = 0.5(k-1)^{(p-1)} - (k-1)^p$ and $Q(s_q, k-1) = R(s_q, k-1) + V_{\pi}(s_{q+1}) = \sum_{q=1}^{p} R(s_q, 0)$. Hence, q \notin $\text{ \bf IS}(\pi)$ for all $q<i$.

\subfile{GPI_smaller}

\subfile{SPI_construct}

%%%%%%%%%%%%%%%%%%%%%%%%%%%%%%%%%%%%%%%%%%%%%%%%%%%%%%%%%%%%%%%%%%%%%%%%%%%%%%%%%
% \section{RANDOM POLICY ITERATION}\label{sec:rpi}
% The above MDP constructed is such that at any time step, only one state is improvable i.e $|{\bf IS}(\pi_t)| = 1$. Furthermore, until the improvable state has the policy set to the best action, it remains to be only state that is improvable. Unlike HPI, as in Random Policy Iteration (RPI), the improving action is selected randomly from ${\bf IA}(\pi_t, {\bf IS}(\pi_t))$, the expected number of iterations required will be scaled by $\log k$ instead of $k$ and so RPI would take $\Omega(n\log k)$ iterations for such an MDP. \doubt{Should we prove this?}
%%%%%%%%%%%%%%%%%%%%%%%%%%%%%%%%%%%%%%%%%%%%%%%%%%%%%%%%%%%%%%%%%%%%%%%%%%%%%%%%%
\section{Simple Policy Iteration}\label{sec:spi}

In Section \ref{sec:api}, we showed a lower bound of $\Omega(k^{n/2})$ iterations for a new, carefully-designed variant of PI, while in Section \ref{sec:hpi}, we provided lower bounds that apply to all PI variants that use index-based or random action-selection. In this section, we investigate the behaviour of Simple PI on multi-action MDPs. Recall that this variant can visit each of the $2^{n}$ policies for an $n$-state, $2$-action MDP \cite{Melekopoglou1994}. Simple PI assumes an arbitrary, fixed indexing of states, and always switches the improvable state with the largest index. We consider index-based and random action selection for $k \geq 3$.

%In Simple Policy Iteration (SPI), the policy of an arbitrary improvable state is switched to an arbitrary improving action. Specifically, the improvable state with the highest index is selected and its policy is switched to the improvable action with the highest index. We prove a $O(k2^{n})$ lower bound for a family of deterministic MDPs which we derive through a simple modification on the 2-action MDPs proposed in \cite{Melekopoglou1994}. 

\subsection{Construction}\label{sec:spi_construction}

Fig.~\ref{fig:spi_construction} shows our construction $H(n, k)$, which generalises the one proposed by Melekopoglou and Condon \cite{Melekopoglou1994}. The MDP  has $n$ non-terminal states, $s_{1}, s_{2}, \dots, s_{n}$, and two terminal states. For $i \in \{1, 2, \dots, n\}$, each state $s_{i}$ has a ``partner" state $s^{\prime}_{i}$, from which two equiprobable outgoing transitions do not depend on action. In principle these states can be removed and the transition probabilities from $s_{1}, s_{2}, \dots, s_{n}$ modified accordingly. Whereas the original construction for $k = 2$ only gives a reward of $-1$ on reaching one of the terminal states, our generalisation also associates rewards with state-action pairs. %Note that if $R(s, a)$ is the reward obtained on taking $a \in A$ from $s \in S$, and additionally $\rho(s^{\prime})$ is the reward given on reaching $s^{\prime} \in S$, we can equivalently administer $R^{\prime}(s, a) = R(s, a) + \sum_{s^{\prime} \in S} T(s, a, s^{\prime}) \rho(s^{\prime})$ when action $a$ is taken from state $s$. The single reward function $R^{\prime}(\cdot, \cdot)$ complies with our definition in Section~\ref{sec:intro}.

As in the original construction, one action, say $0$, transitions deterministically, with no reward, from each state $s_{i}$ to state $s_{i - 1}$ for $i \in \{2, 3, \dots, n\}$, and from $s_{1}$ to a terminal state. We design the other actions $j \in \{1, 2, \dots, k - 1\}$ from each state to transition deterministically to the corresponding partner state; action $j$ gets reward $\epsilon / 2^{k - 1 - j}$, where $\epsilon = 2^{-n}$.

%Taking $\epsilon = 2^{-n}$ ensures that the sequence of policies visited for $k = 2$ case does not change from the original construction (in which $\epsilon = 0$).
%$\epsilon$ is chosen so that the optimal policy for the MDP is $0\dots 01$. Note that under the optimal policy value of a partner state $s_{n}'\geq 2$ is $-2^{-(s_{n}'-1)}+\frac{R_{k-1}}{2}$. A necessary condition so that the optimal policy is realized is $Q(s_{n}',k-1)<Q(s_{n}',0) \: \forall s_{n}'>1$. This is equivalently $R_{k-1} > R_{k-1} -2^{-(i-1)}+\frac{R_{k-1}}{2} \: \forall i>1 $. Therefore $R_{k-1} = \epsilon < 4\times 2^{-n}$. Hence we choose $\epsilon = 2^{-n}$ \shivaram{I believe $\epsilon$ must depend on $n$, maybe also $k$. No?} \reply{Have formulated $\epsilon$} 
%Our generalised setup ensures that if $s$ is an improvable state currently taking action $a \in \{0, 1, \dots, k - 1\}$, then actions $a + 1, a + 2, \dots, k - 1$ are all improving actions.

%The family of MDPs maintain a similar structure to the family proposed in \cite{Melekopoglou1994} with average, sink and state vertices. We augment this family with deterministic transitions from a state vertex s to its average vertex s' for $k>2$ actions with reward $R_{k}$ for action k $\geq 1$. A higher index action is assigned a lower reward i.e. $R_{i}\leq R_{i-1}  \forall i \in [2,k-1]$. The construction has been shown in  Fig ~\ref{fig:spi_construction}. We set $R_{i} = \epsilon \times 2^{-(i+1)} $ where $\epsilon = 10^{-4}$. 

\subsection{Lower Bounds}\label{sec:spi_lb}

While the original construction uses $\epsilon = 0$, we require the rewards on the actions to be different so that more policies can be visited by PI. Our generalised setup ensures that if $s$ is an improvable state currently taking action $a \in \{0, 1, \dots, k - 2\}$, then actions $a + 1, a + 2, \dots, k - 1$ are all improving actions. Moreover, taking $\epsilon = 2^{-n}$ retains the structure of the trajectory taken by Simple PI on $H(n, 2)$. Indeed for $t \in \{1, 2, 3, \dots, 2^{n}\}$, if $\pi_{t} \in \{0, 1\}^{n}$ is the $t$-th policy visited on $H(n, 2)$, then $\pi^{\prime}_{t} \in \{0, k - 1\}^{n}$, which has every occurrence of $1$ replaced by $k - 1$ in $\pi_{t}$, is the $t$-th policy from $\{0, k - 1\}^{n}$ visited on $H(n, k)$. 

The reason we get scaling of lower bound with $k$ is that corresponding to every switch from action 0 to action 1 on $H(n, 2)$, there is a progression through $k - 1$ actions---$0, 1, \dots, k - 1$---on 
$H(n, k)$, if using index-based action selection. With random action selection $\theta(\log(k))$ actions are visited in expectation,  following the reasoning given in Section~\ref{sec:hpi}. Since Simple PI makes $\Omega(2^{n})$ switches from action $0$ to action $1$ on $H(n, 2)$, we can generalise as below.

\begin{theorem}[Simple PI Lower Bounds]
On $H(n, k)$, if initialised with policy $0^{n}$, Simple PI takes $\Omega(k\cdot2^{n})$ iterations with index-based action selection, and $\Omega(\log(k)\cdot2^{n})$ iterations in expectation with random action selection.
\end{theorem}

%The proof follows from the 2-action MDP proof given in \cite{Melekopoglou1994}. Let their MDP construction be denoted by $\mathbb{F}$. 
% $$F \mathrm{F} \mathit{F} \mathscr{F}\mathcal{F} \mathbb{F} \mathfrak{F} \textswab{F} \textgoth{F}$$

%Every switch from action 0 to action 1 in $\mathbb{F}$ takes k iterations in our construction. This is because all actions k$\geq 1$ are improvable for a state that iteratively switches from action 0 to action 1. Due to the reward assignment in our construction, action k-1 is the best improvable action and action 1 is the worst improvable action. Since SPI chooses the lowest indexed improvable action at every iteration, our construction forces k iterations to switch from 0 to k-1.When SPI is run on $\mathbb{F}$, more than half the switching is from action 0 to action k-1. Since a $O(2^n)$ lower bound for a 2-action MDP is proved in \cite{Melekopoglou1994}, our construction of a k-action,n-state MDP gives a lower bound of $O(k2^n)$. 
%\par In the case of RSPI, as the improving action is selected uniform-randomly from $\bf IA$, the lower bound for it would instead be $O(2^n\log k)$ \doubt{can we refer to the lemma from section 5 regarding $\log(k)$ steps?}
%%%%%%%%%%%%%%%%%%%%%%%%%%%%%%%%%%%%%%%%%%%%%%%%%%%%%%%%%%%%%%%%%%%%%%%%%%%%%%%%%
% \section{RANDOMIZED SIMPLE POLICY ITERATION}\label{sec:rspi}
% We demonstrate
% Similar to the argument for RPI, 
%%%%%%%%%%%%%%%%%%%%%%%%%%%%%%%%%%%%%%%%%%%%%%%%%%%%%%%%%%%%%%%%%%%%%%%%%%%%%%%%
\section{Conclusion and Future Work}\label{sec:conclusions}

PI~\cite{howard:dp} is a widely-used family of algorithms for solving MDPs, which model sequential decision making tasks in stochastic domains. While there is a fair amount of work on the theoretical analysis of PI, the literature on lower bounds is relatively sparse. In particular, existing lower bounds on the running-time of PI on $n$-state, $k$-action MDPs either assume $k = 2$ or take $k$ to be dependent on $n$. We present the first non-trivial lower bounds for the general case of $k \geq 2$.

We consider a deterministic, index-based action-selection strategy, as well as a randomised one. When coupled with Simple PI~\cite{Melekopoglou1994}---earlier analysed for $k = 2$---these strategies increase the corresponding lower bound by factors of $k$ and $\log(k)$, respectively. We also show the same scaling in terms of $k$ for the tightest lower bound known yet for Howard's PI on $2$-action MDPs. Indeed the resulting lower bounds of $\Omega(kn)$ and $\Omega(\log(k) \cdot n)$ iterations apply to all PI variants that use index-based and random action-switching, respectively. Our constructions do not yield non-trivial lower bounds when used in conjunction with the popular ``max-Q'' action selection strategy, which needs further investigation.

From a lower-bounding perspective, the major open question is whether there is an $n$-state, $k$-action MDP on which some variant of PI can visit all of the $k^{n}$ policies. While the answer is affirmative for $k = 2$~\cite{Melekopoglou1994}, we are yet unaware what it is for $k \geq 3$. The tightest lower bound we show in this paper is $\Omega(k^{n/2})$ iterations, which is significant in having $\sqrt{k}$, rather than a constant, in the base of the exponent. Future work could explore improvements to our lower bound. Another possibility is to show an upper bound smaller than $k^{n}$ that simultaneously holds for all PI variants in the case of $k \geq 3$.

\section*{Acknowledgements}

Shivaram Kalyanakrishnan was partially supported by SERB grant ECR/2017/002479.

%We have thus devised an adversarial policy iteration algorithm for which a lower bound of $O(k^{0.5n})$ was achieved. This is still far from the $k^n$ trivial upper bound and further investigation has to be done to see there exists a PI algorithm that has a lower bound of $O(k^n)$. We have achieved strong lower bounds for HPI, RPI, SPI, and RSPI for $k(\geq 2)$-action MDPs of the form $O(kn)$, $O(\log kn)$, $O(k2^n)$ and $O(2^n \log k)$ respectively. It remains to see if it is possible to construct an family of MDPs through which these PI algorithms achieve lower bounds with a stronger dependence on k. 

%%%%%%%%%%%%%%%%%%%%%%%%%%%%%%%%%%%%%%%%%%%%%%%%%%%%%%%%%%%%%%%%%%%%%%%%%%%%%%%%
% \section{ACKNOWLEDGMENTS}

% The authors gratefully acknowledge the contribution of National Research Organization and reviewers' comments.
% \doubt{Taken directly from sample tex, what do we write here?}

%%%%%%%%%%%%%%%%%%%%%%%%%%%%%%%%%%%%%%%%%%%%%%%%%%%%%%%%%%%%%%%%%%%%%%%%%%%%%%%%

\bibliographystyle{unsrt}
\bibliography{references}
           

\newpage
  \appendix

\subsection{Trajectory of Peculiar PI on $F(3, 3)$}
\label{appendix:f33trajectory}

Below we list the sequence of trajectories visited by Peculiar PI (the PI variant from Section~\ref{sec:api}) on $F(3, 3)$, when initialised with policy $0^{3} \cdot 0^{3}$. Each line begins with a ``balanced'' policy (of the form $x \cdot x$ for $x \in \{0, 1, 2\}^{3}$); to its right is the sequence of policies taken to reach the next balanced policy.

% \addtolength{\textheight}{-7cm}   % This command serves to balance the column lengths
%                                   % the textheight of the last page by a suitable amount.
%                                   % This command does not take effect until the next page
%                                   % so it should come on the page before the last. Make
%   

%\begin{table}[htbp]
%  \centering
%  \caption{Trajectory for $\family{F}(3,3)$}
\begin{center}
\footnotesize 
%    \begin{tabular}{|c||c||c||c||c||c|}
    \begin{tabular}{|cccccc|}
    \hline
    $\bm{0 0 0 \cdot 0 0 0}$ & $0 0 0 \cdot 0 0 1$ &       &       &       &  \\
    $\bm{0 0 1 \cdot 0 0 1}$ & $0 0 1 \cdot 0 0 2$ &       &       &       &  \\
    $\bm{0 0 2 \cdot 0 0 2}$ & $0 0 2 \cdot 0 1 2$ & $0 0 2 \cdot 0 1 0$ & $0 0 0 \cdot 0 1 0$ &       &  \\
    $\bm{0 1 0 \cdot 0 1 0}$ & $0 1 0 \cdot 0 1 1$ &       &       &       &  \\
    $\bm{0 1 1 \cdot 0 1 1}$ & $0 1 1 \cdot 0 1 2$ &       &       &       &  \\
    $\bm{0 1 2 \cdot 0 1 2}$ & $0 1 2 \cdot 0 2 2$ & $0 1 2 \cdot 0 2 0$ & $0 1 0 \cdot 0 2 0$ &       &  \\
    $\bm{0 2 0 \cdot 0 2 0}$ & $0 2 0 \cdot 0 2 1$ &       &       &       &  \\
    $\bm{0 2 1 \cdot 0 2 1}$ & $0 2 1 \cdot 0 2 2$ &       &       &       &  \\
    $\bm{0 2 2 \cdot 0 2 2}$ & $0 2 2 \cdot 1 2 2$ & $0 2 2 \cdot 1 0 2$ & $0 2 2 \cdot 1 0 0$ & $0 2 0 \cdot 1 0 0$ & $0 0 0 \cdot 1 0 0$ \\
    $\bm{1 0 0 \cdot 1 0 0}$ & $1 0 0 \cdot 1 0 1$ &       &       &       &  \\
    $\bm{1 0 1 \cdot 1 0 1}$ & $1 0 1 \cdot 1 0 2$ &       &       &       &  \\
    $\bm{1 0 2 \cdot 1 0 2}$ & $1 0 2 \cdot 1 1 2$ & $1 0 2 \cdot 1 1 0$ & $1 0 0 \cdot 1 1 0$ &       &  \\
    $\bm{1 1 0 \cdot 1 1 0}$ & $1 1 0 \cdot 1 1 1$ &       &       &       &  \\
    $\bm{1 1 1 \cdot 1 1 1}$ & $1 1 1 \cdot 1 1 2$ &       &       &       &  \\
    $\bm{1 1 2 \cdot 1 1 2}$ & $1 1 2 \cdot 1 2 2$ & $1 1 2 \cdot 1 2 0$ & $1 1 0 \cdot 1 2 0$ &       &  \\
    $\bm{1 2 0 \cdot 1 2 0}$ & $1 2 0 \cdot 1 2 1$ &       &       &       &  \\
    $\bm{1 2 1 \cdot 1 2 1}$ & $1 2 1 \cdot 1 2 2$ &       &       &       &  \\
    $\bm{1 2 2 \cdot 1 2 2}$ & $1 2 2 \cdot 2 2 2$ & $1 2 2 \cdot 2 0 2$ & $1 2 2 \cdot 2 0 0$ & $1 2 0 \cdot 2 0 0$ & $1 0 0 \cdot 2 0 0$ \\
    $\bm{2 0 0 \cdot 2 0 0}$ & $2 0 0 \cdot 2 0 1$ &       &       &       &  \\
    $\bm{2 0 1 \cdot 2 0 1}$ & $2 0 1 \cdot 2 0 2$ &       &       &       &  \\
    $\bm{2 0 2 \cdot 2 0 2}$ & $2 0 2 \cdot 2 1 2$ & $2 0 2 \cdot 2 1 0$ & $2 0 0 \cdot 2 1 0$ &       &  \\
    $\bm{2 1 0 \cdot 2 1 0}$ & $2 1 0 \cdot 2 1 1$ &       &       &       &  \\
    $\bm{2 1 1 \cdot 2 1 1}$ & $2 1 1 \cdot 2 1 2$ &       &       &       &  \\
    $\bm{2 1 2 \cdot 2 1 2}$ & $2 1 2 \cdot 2 2 2$ & $2 1 2 \cdot 2 2 0$ & $2 1 0 \cdot 2 2 0$ &       &  \\
    $\bm{2 2 0 \cdot 2 2 0}$ & $2 2 0 \cdot 2 2 1$ &       &       &       &  \\
    $\bm{2 2 1 \cdot 2 2 1}$ & $2 2 1 \cdot 2 2 2$ &       &       &       &  \\
    $\bm{2 2 2 \cdot 2 2 2}$ &       &       &       &       &  \\
    \hline
    \end{tabular}
\normalsize    
\end{center}

%  \label{tab:traj_6_3}
%\end{table}

For $F(m, k)$, the exact number of policies that are visited using our construction is
$\frac{2k}{k - 1}(k^{m} - 1) -2m + 1$, seen here to be $73$ for $F(3, 3)$.

\subsection{Memoryless Encoding of Peculiar PI}
\label{appendix:memory_less_algo}

The idea of our construction in Section~\ref{sec:api} is to proceed from one balanced policy to the next through a sequence of policy improvement steps. Below we provide a memoryless specification of Peculiar PI, the variant we have designed for this purpose. Given an arbitrary policy of the form $x \cdot y$, where $x, y \in A^{m}$, Peculiar PI identifies the state to switch, denoted $\bar{s} \in S$, as follows.
\begin{center}
\fbox{
  \begin{minipage}{0.45\textwidth}
Define $d  = [y] - [x]$ and if $d \geq 1$, define $b = \lfloor \log_{k}(d) \rfloor.$\\
If $d < 0$: //Cannot arise on $F(m, k)$,  starting from $0^{m} \cdot 0^{m}$.\\
\phantom{aaaa}Set $\bar{s}$ to be an arbitrary state.\\
Else if $d = 0$:\\
\phantom{aaaa}$\bar{s} \gets s^{\prime}_{I(x)}$.\\
Else if $d = 1$:\\
\phantom{aaaa}$\bar{s} \gets s_{m}$.\\
Else if $y_{m} = k - 1$:\\
\phantom{aaaa}$\bar{s} \gets s^{\prime}_{m - b + 1}$. \\
Else:\\
\phantom{aaaa}$\bar{s} \gets s_{m - b}$.
 \end{minipage}
}
\end{center}

First, note that $\bar{s}$ is \textit{not} guaranteed to be an improvable state on every MDP. In fact, it might not be improvable even for $F(m, k)$ for some policies $x \cdot y$. However, if Peculiar PI is initialised with policy $0^{m} \cdot 0^{m}$ on $F(m, k)$, then the procedure outlined here will exactly simulate the trajectory of policies described in the proof of 
 Lemma~\ref{lem:segments}. This property suffices for the purpose of our lower bound. We allow Peculiar PI to be defined arbitrarily when $\bar{s}$ is not an improvable state, or when it does not have the desired choice of improving action (specified next).

If $\bar{s}$ is indeed improvable and $x \cdot y (\bar{s}) \neq k - 1$, then Peculiar PI switches the action $j$ for $\bar{s}$ to $j + 1$ (if $j + 1$ is an improving action). If $\bar{s}$ is improvable and $x \cdot y (\bar{s}) = k - 1$, then 
Peculiar PI switches the action for $\bar{s}$ to $0$ (if $0$ is an improving action). In summary, if $\bar{s}$ is an improvable state and $(x.y(\bar{s}) + 1) \mod k$ is an improving action, Peculiar PI switches to this action. 

%he action for improvable state $\bar{s}$ to $(x.y(\bar{s}) + 1) \mod k$.

%If not, Peculiar PI can perform an arbitrary switch. \doubt{When action is k-1, only improvable action is 0. thus $a = (x.y(\bar{s}) + 1)\mod k$}

%On other MDPs, or if initialised with certain policies even on $F(m, k)$, it could happen that $\bar{s}$ is not improvable, or it is associated with action $k - 1$. Since we are only interested in a \textit{lower} bound, we ignore the behaviour of Peculiar PI in such cases.

Observe that the procedure outlined above can be implemented using $\text{poly}(n, k)$ arithmetic operations and space.

\subsection{Extending Lower Bounds to Discounted Reward Setting}\label{appendix:total_to_discounted}

All three of our MDP families---$F(m, k)$ (Section~\ref{sec:api}), $G(m, k)$ (Section~\ref{sec:hpi}), and $H(m, k)$ (Section~\ref{sec:spi})---are defined under the total reward setting. To generalise our lower bounds to the discounted reward setting, we begin by observing that for each MDP family, the following  properties are satisfied.
\begin{enumerate}
\item For all $\pi:S \to A$, $s \in S$, and $a, a^{\prime} \in A$: $$(a \neq a^{\prime}) \Longrightarrow Q^{\pi}(s, a) \neq Q^{\pi}(s, a^{\prime}).$$
\item There is a finite number $L$ such that starting 
from any state, taking any actions, the number of steps to termination is at most $L$.
\item Rewards are all bounded; assume they lie in $[-R_{\max}, R_{\max}]$ for finite $R_{max} > 0$.

\end{enumerate}
Define $$\Delta \eqdef \min_{\pi:S \to A, s \in S, a, a^{\prime} \in A, a \neq a^{\prime}} |Q^{\pi}(s, a) - Q^{\pi}(s, a^{\prime})|.$$ Since the first property is satisfied, we have $\Delta > 0$. The second property implies that every $Q$-value may be written as a sum of $L$ (expected) rewards: $$Q = X_{1} + X_{2} + X_{3} + \dots + X_{L}.$$ Now, if we use a discount factor $\gamma \in [0, 1]$, we have
$$Q_{\gamma} = X_{1} +  \gamma X_{2} + \gamma^{2} X_{3} + \dots + \gamma^{L - 1} X_{L}.$$ Consequently, we have 
\begin{align*}
|Q^{\pi}(s, a) - Q^{\pi}_{\gamma}(s, a)|
&= |\sum_{i = 2}^{L} (1 - \gamma^{i - 1}) X_{i}|\\
&\leq |\sum_{i = 2}^{L} (1 - \gamma^{i - 1}) R_{\max}|\\
&\leq (L - 1)(1 - \gamma^{L - 1})R_{\max}.
\end{align*}
For $\gamma > \gamma_{0} = \left(\max\{1 - \frac{\Delta}{2(L - 1)R_{\max}}, 0\}\right)^{\frac{1}{L - 1}}$, we observe that $|Q^{\pi}(s, a) - Q^{\pi}_{\gamma}(s, a)| < \frac{\Delta}{2}$. Hence, for all $\gamma \in (\gamma_{0}, 1]$, the relative order of $Q_{\gamma}$ values is identical for all policies, states, and actions.

The lower bounds we have provided are all for PI variants that are defined solely based on the relative order among Q-values for each state and action. Consequently these algorithms follow the same trajectories for all 
$\gamma \in (\gamma_{0}, 1]$.

\subsection{Proof of Lemma~\ref{lemma:hpi_is_ia}}
\label{sec:hpilemmaproof}

Recall that for $i \in \{1, 2, \dots, n\}$, $j \in \{0, 1, \dots, k - 2\}$ we have
$\pi_{ij} = 0^{i - 1}j(k - 1)^{n - i}.$ For $u \in \{1, 2, \dots, n\}$, we observe
\begin{center}
    $V^{\pi_{ij}}(s_u)= 
\begin{cases}
    -2^{u} &\text{ if } u < i,\\
   -2^{i}\left(\frac{1}{2} + \frac{k-j}{2k}\right) &\text{ if } u = i,\\
    0  &\text{ if } u>i.
\end{cases}$

% \textcolor{red}{Substitute for the actual value of  of $R(s_{i}, j)$ here; I don't think we have clearly used/defined ``$R(s_i, j)$'' for this class of MDPs in the main text.}

\end{center}

In order to prove that 
$\text{\bf IS}(\pi_{ij}) = \{i\}$ and $\text{\bf IA}(\pi_{ij}, i) = \{j + 1, j + 2, \dots, k - 1\}$, first we show that
$i \in \text{\bf IS}(\pi_{ij})$ and $\{j+1, j+2, \dots, k-1\} \subset \text{\bf IA}(\pi_{ij}, i)$. Observe that for $j' \in \{j + 1, j + 2, \dots, k - 2\}$, \begin{align*}
    V^{\pi_{ij}}(s_i) &= -2^{i}\left(\frac{1}{2} + \frac{k-j}{2k}\right) < -2^{i}\left(\frac{1}{2} + \frac{k-j'}{2k}\right) \\
    &=Q^{\pi_{ij}}(s_i, j'),
\end{align*}
%Hence, $i \in \text{\bf IS}(\pi)$ and $\{j+1, j+2, \dots, k-2\} \subset \text{\bf IA}(\pi, i)$.
and also, $V^{\pi_{ij}}(s_i) < 0 = Q^{\pi_{ij}}(s_i, k - 1)$. Hence, $i \in \text{\bf IS}(\pi_{ij})$ and $\{j+1, j+2, \dots, k-1\} \subset \text{\bf IA}(\pi_{ij}, i)$.

%We further show that $u \not\in \text{\bf IS}(\pi)$ for $u \neq i$. 

%\noindent \underline{$u = i$.} 

%In this case f

% \textcolor{red}{Write both inequalities above from right to left, and equate the last term with $V^{\pi_{ij}}(s_{i})$.} 

Next, we show that $u \notin \text{\bf IS}(\pi_{ij})$
for $u \in \{1, 2, \dots, i - 1\} \cup \{i + 1, i + 2, \dots, n\}$ by considering separate cases.\\
% \textcolor{red}{Place the material below under the corresponding cases.}

\noindent \underline{$u \in \{1, 2, \dots, i - 2\}$.} In this case, for $j' \in \{1, 2, \dots, k-2\}$,
\begin{align*}
    Q^{\pi_{ij}}(s_u, j') = -2^{u}\left(\frac{1}{2} + \frac{k-j'}{2k}\right) -2^{u+1}\left(\frac{1}{2} - \frac{k-j'}{2k}\right),
\end{align*}
and $Q^{\pi_{ij}}(s_u, k - 1) = -2^{u+1}$. Thus, for $j' \in \{1, 2, \dots, k-1\}$, $Q^{\pi_{ij}}(s_u, j') < -2^{u} = V^{\pi_{ij}}(s_u)$.\\ 

\noindent \underline{$u = i - 1$.} In this case, for $j' \in \{ 1, 2,  \dots, k-2\}$,
\begin{align*}
    Q^{\pi_{ij}}(s_u, j') = -2^{u}\left(\frac{1}{2} + \frac{k-j'}{2k}\right) + V^{\pi_{ij}}(s_i)\left(\frac{1}{2} - \frac{k-j'}{2k}\right),
\end{align*}
and $Q^{\pi_{ij}}(s_u, k - 1) = V^{\pi_{ij}}(s_i)$. Substituting for $V^{\pi_{ij}}(s_i)$, we get, for $j' \in \{1, 2, \dots, k-1\}$, $Q^{\pi_{ij}}(s_u, j') < V^{\pi_{ij}}(s_u)$.\\

\noindent \underline{$u \in \{i + 1, i + 2, \dots, n\}$.} In this case for $j' \in \{0, 1, \dots, k-2\}$,
\begin{align*}
    Q^{\pi_{ij}}(s_u, j') = -2^{u}\left(\frac{1}{2} + \frac{k-j'}{2k} \right) < 0 = V^{\pi_{ij}}(s_u).
\end{align*}

\end{document}

%% file: API_construct_inv.tex
\begin{figure*}[t!]
\centering
\begin{adjustbox}{max width=0.81\textwidth}
    \centering
        \begin{tikzpicture}[node distance=3.8cm,auto,transform shape,-{Latex[length=1.8mm,width=1mm]},>=Latex]
    \tikzset{state/.style={circle,draw=black,minimum size=10mm}}

    \node[state]             (s0)       {$s_{m}$};
    \node[state,below of=s0] (s0_prime) {$s^{\prime}_{m}$};
    \node[state,left of=s0] (s1)       {$s_{m - 1}$};
    \node[state,below of=s1] (s1_prime) {$s^{\prime}_{m - 1}$};
    \node[state,left of=s1] (s2)       {$s_{m - 2}$};
    \node[state,below of=s2] (s2_prime) {$s^{\prime}_{m - 2}$};
    \node[left=1cm of s2] (dots1)       {\dots};
    \node[left=1cm of s2_prime] (dots2) {\dots};
    \node[state,left=1cm of dots1] (sn-1) {$s_{2}$};
    \node[state,below of=sn-1] (sn-1_prime)  {$s^{\prime}_{2}$};
    \node[state,left of=sn-1] (sn) {$s_{1}$};
    \node[state,left of=sn-1_prime] (sn_prime)  {$s^{\prime}_{1}$};
    \node[state,accepting,below left=2cm of sn] (term) {$s_T$};
    %% ----------------------
    \draw
        (s0) edge [black,bend right=50, looseness=1.3] node[above] {\small{$1:1$}}(s1)
        (s0) edge [black, bend right=30, looseness=1] node[above] {\small{$2:2$}} node[below=-0.1] {\vdots} (s1)
        (s0) edge [black] node[below]{\small{$k-1:k-1$}}  (s1)
        
        (s0_prime) edge [black, bend right=10, looseness=1] node[above,sloped] {\small{$1:1$}}  (s1)
        (s0_prime) edge [black, bend left=5] node[above,sloped] {\small{$2:2$}} node[below=-0.1,sloped] {\vdots}  (s1)
        (s0_prime) edge [black, bend left=25] node[below,sloped] {\small{$k-1:k-1$}} (s1)
        
        (s0) edge [black, bend left=48] node[below,pos=0.2,sloped] {\small{$0:0$}}  (s1_prime)
        (s0_prime) edge [black] node[below] {\small{$0:0$}}  (s1_prime);
    %% -----------------------
    \draw
        (s1) edge [black,bend right=50, looseness=1.3] node[above] {\small{$1:k$}}(s2)
        (s1) edge [black, bend right=30, looseness=1] node[above] {\small{$2:2k$}} node[below=-0.1] {\vdots} (s2)
        (s1) edge [black] node[below]{\small{$k-1:(k-1)k$}}  (s2)
        
        (s1_prime) edge [black, bend right=10] node[above,sloped] {\small{$1:k$}}  (s2)
        (s1_prime) edge [black, bend left=5] node[above,sloped] {\small{$2:2k$}} node[below=-0.1,sloped] {\vdots}  (s2)
        (s1_prime) edge [black, bend left=25] node[below,sloped] {\small{$k-1:(k-1)k$}} (s2)
        
        (s1) edge [black, bend left=48] node[below,pos=0.2,sloped] {\small{$0:0$}}  (s2_prime)
        (s1_prime) edge [black] node[below] {\small{$0:0$}}  (s2_prime);
    
    \draw
        (sn-1) edge [black,bend right=50, looseness=1.2] node[above] {\small{$1:k^{m-2}$}}(sn)
        (sn-1) edge [black, bend right=30, looseness=1] node[above] {\small{$2:2k^{m-2}$}} node[below=-0.1] {\vdots} (sn)
        (sn-1) edge [black] node[below]{\small{$k-1:k^{m-2}(k-1)$}}  (sn)
        
        (sn-1_prime) edge [black,bend right=10] node[above,sloped] {\small{$1:k^{m-2}$}}  (sn)
        (sn-1_prime) edge [black,bend left=5] node[above,sloped] {\small{$2:2k^{m-2}$}} node[below=-0.2,sloped] {\vdots}  (sn)
        (sn-1_prime) edge [black, bend left=25] node[below,sloped] {\small{$k-1:(k-1)k^{m-2}$}} (sn)
        
        (sn-1) edge [black, bend left=48] node[below,pos=0.2,sloped] {\small{$0:0$}}  (sn_prime)
        (sn-1_prime) edge [black] node[below] {\small{$0:0$}}  (sn_prime);    
    
    %% -----------------------
    \draw
        (sn) edge [black,bend right=40,looseness=1.3] node[above,sloped] {\small{${k-1}:k^{m - 1}(k-1)$}}(term)
        (sn) edge [black,bend right=20] node[below=-0.1,sloped] {\small{$1:k^{m - 1}$}} node[above,sloped] {\vdots} (term)
        (sn) edge [black] node[below,sloped]{\small{$0:0$}}  (term)
        
        (sn_prime) edge [black,bend left=40,looseness=1.4] node[below,sloped] {\small{$k-1:k^{m - 1}(k-1)$}}(term)
        (sn_prime) edge [black,bend left=20] node[above=-0.1,sloped] {\small{$1:k^{m - 1}$}} node[below=-0.2,sloped] {\vdots} (term)
        (sn_prime) edge [black] node[above,sloped]{\small{$0:0$}}  (term);

    \normalsize

  \end{tikzpicture}
\end{adjustbox}
\vspace{-0.4cm}
\caption{The deterministic MDP $F(m, k)$ with $2m$ non-terminal states, a single terminal state $s_{T}$, and $k$ actions. States $s_{1}, s_{2}, \dots, s_{m}$ implement a $k$-ary ``counter''; each has an associated partner state. Each edge, labeled ``action: reward'' represents the corresponding transitions. No discounting is used.} 
\vspace{-0.4cm}
\label{fig:adversarial_family2}
\end{figure*}
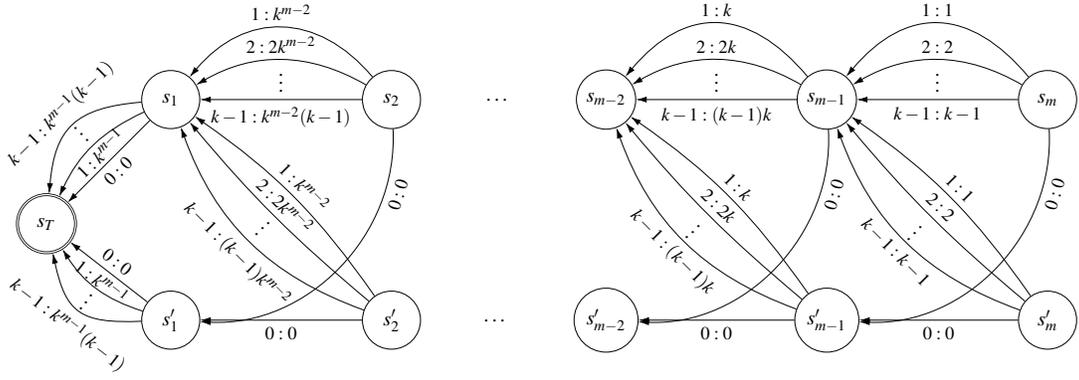

%% file: GPI_smaller.tex
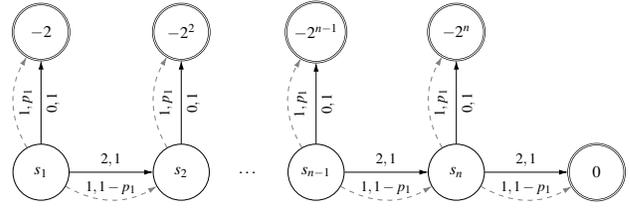
\begin{figure}[t]
\centering
\begin{adjustbox}{max width=0.95\linewidth}
    \centering
    \begin{tikzpicture}[node distance=3cm,auto,transform shape,-{Latex[length=1.8mm,width=1mm]},>=Latex]
    \tikzset{state/.style={circle,draw=black,minimum size=12mm}}
    
    % \node[]                  (s0) {};
    \node[state] (s1) {$s_1$};
    \node[state,accepting,above of=s1] (s1_term) {$-2$};
    \node[state,right of=s1] (s2) {$s_2$};
    \node[state,accepting,above of=s2] (s2_term) {$-2^2$};
    \node[right=0.5cm of s2] (dots) {\dots};
    \node[state, right=0.5cm of dots] (sn-1) {$s_{n-1}$};
    \node[state,accepting,above of=sn-1] (sn-1_term) {$-2^{n-1}$};
    \node[state,right of=sn-1] (sn) {$s_n$};
    \node[state,accepting,above of=sn] (sn_term) {$-2^n$};
    \node[state,right of=sn,accepting] (term) {$0$};

    %% ----------------------
    \draw
        (s1) edge [black,sloped] node[below]{\small{$0,1$}}(s1_term)
        (s1) edge [gray,bend left=30,dashed,sloped] node[black,below,rotate=180]{\small{$1,p_1$}}(s1_term)
        
        (s1) edge [black] node[above] {\small{$2,1$}}(s2)
        (s1) edge [gray,bend right,dashed] node[black,above]{\small{$1,1-p_1$}}(s2)
        
        (s2) edge [black,sloped] node[below]{\small{$0,1$}}(s2_term)
        (s2) edge [gray,bend left=30,sloped,dashed] node[black,below,rotate=180]{\small{$1,p_1$}}(s2_term)
    
        (sn-1) edge [black,sloped] node[below]{\small{$0,1$}}(sn-1_term)
        (sn-1) edge [gray,bend left=30,sloped,dashed] node[black,below,rotate=180]{\small{$1,p_1$}}(sn-1_term)
        
        (sn-1) edge [black] node[above] {\small{$2,1$}}(sn)
        (sn-1) edge [gray,bend right,dashed] node[black,above]{\small{$1,1-p_1$}} (sn)
        
        (sn) edge [black,sloped] node[below]{\small{$0,1$}}(sn_term)
        (sn) edge [gray,bend left=30,sloped,dashed] node[black,below,rotate=180]{\small{$1,p_1$}}(sn_term)
        
        (sn) edge [black] node[above] {\small{$2,1$}}(term)
        (sn) edge [gray,bend right,dashed] node[black,above]{\small{$1,1-p_1$}}(term)
        ;
    %% -----------------------

    \normalsize

  \end{tikzpicture}
\end{adjustbox} 
\vspace{-0.2cm}
    \caption{The stochastic MDP $G(n, 3)$, used to illustrate the structure of $G(n, k)$. Labels on arrows mark ``action, probability''; terminal states show  rewards.
    While actions $0$ and $k - 1$ are deterministic, all others are stochastic, with transition probabilities as specified in Section~\ref{sec:hpi_construction}.}
    \label{fig:gpi_construction}
\vspace{-0.6cm}
\end{figure}

%% file: SPI_construct.tex
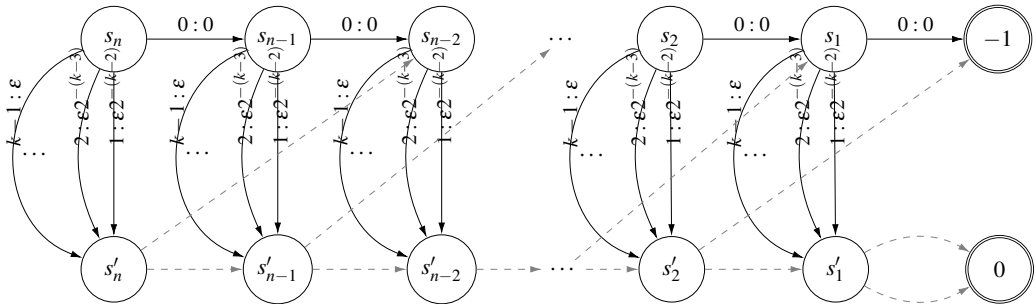
\begin{figure*}[ht]
\centering
\begin{adjustbox}{max width=0.78\linewidth}
    \centering
    \begin{tikzpicture}[node distance=2.5cm,auto,transform shape,-{Latex[length=1.8mm,width=1mm]},>=Latex]
    \tikzset{state/.style={circle,draw=black,minimum size=10mm}}

    \node[state]             (sn)       {$s_n$};
    \node[state,below=2.5cm of sn] (sn_prime) {$s_n'$};
    \node[state,right of=sn] (s_n-1)       {$s_{n-1}$};
    \node[state,right of=sn_prime] (s_n-1_prime) {$s_{n-1}'$};
    \node[state,right of=s_n-1] (s_n-2)       {$s_{n-2}$};
    \node[state,right of=s_n-1_prime] (s_n-2_prime) {$s_{n-2}'$};
    \node[right=1cm of s_n-2] (dots1)       {\dots};
    \node[right=1cm of s_n-2_prime] (dots2) {\dots};
    \node[state,right=0.8cm of dots1] (s2)       {$s_{2}$};
    \node[state,right=0.8cm of dots2] (s2_prime) {$s_{2}'$};
    \node[state,right of=s2] (s1)       {$s_{1}$};
    \node[state,right of=s2_prime] (s1_prime) {$s_{1}'$};
    \node[state,right of=s1,accepting] (term1)       {$-1$};
    \node[state,right of=s1_prime, accepting] (term2) {$0$};
    %% ----------------------
    \draw
        (sn) edge [black] node[] {\small{$0:0$}}(s_n-1)
        (s_n-1) edge [black] node[] {\small{$0:0$}}(s_n-2)
        (s2) edge [black] node[] {\small{$0:0$}}(s1)
        (s1) edge [black] node[] {\small{$0:0$}}(term1)
        
        (sn_prime) edge [gray,dashed] node[] {}(s_n-1_prime)
        (s_n-1_prime) edge [gray,dashed] node[] {}(s_n-2_prime)
        (s_n-2_prime) edge [gray,dashed] node[] {}(dots2)
        (s2_prime) edge [gray,dashed] node[] {}(s1_prime)
        
        (sn) edge [black] node[sloped,rotate=180] {\small{$1:\epsilon 2^{-(k-2)}$}}(sn_prime)
        (sn) edge [black, bend right=25] node[sloped,rotate=180] {\small{$2:\epsilon2^{-(k-3)}$}}(sn_prime)
        (sn) edge [black, bend right=70,looseness=1.2] node[sloped,rotate=180] {\small{$k-1:\epsilon$}} node[] {\dots} (sn_prime)
        
        (s_n-1) edge [black] node[sloped, rotate=180] {\small{$1:\epsilon 2^{-(k-2)}$}}(s_n-1_prime)
        (s_n-1) edge [black, bend right=25] node[sloped] {\small{$2:\epsilon 2^{-(k-3)}$}}(s_n-1_prime)
        (s_n-1) edge [black, bend right=70,looseness=1.2] node[sloped] {\small{$k-1:\epsilon$}} node[] {\dots} (s_n-1_prime)
        
        (s_n-2) edge [black] node[sloped, rotate=180] {\small{$1:\epsilon 2^{-(k-2)}$}}(s_n-2_prime)
        (s_n-2) edge [black, bend right=25] node[sloped] {\small{$2:\epsilon 2^{-(k-3)}$}}(s_n-2_prime)
        (s_n-2) edge [black, bend right=70,looseness=1.2] node[sloped] {\small{$k-1:\epsilon$}} node[] {\dots} (s_n-2_prime)
        
        (s2) edge [black] node[sloped, rotate=180] {\small{$1:\epsilon 2^{-(k-2)}$}}(s2_prime)
        (s2) edge [black, bend right=25] node[sloped,rotate=180] {\small{$2:\epsilon 2^{-(k-3)}$}}(s2_prime)
        (s2) edge [black, bend right=70,looseness=1.2] node[sloped,rotate=180] {\small{$k-1:\epsilon $}} node[] {\dots} (s2_prime)
        
        (s1) edge [black] node[sloped, rotate=180] {\small{$1:\epsilon 2^{-(k-2)}$}}(s1_prime)
        (s1) edge [black, bend right=25] node[sloped,rotate=180] {\small{$2:\epsilon 2^{-(k-3)}$}}(s1_prime)
        (s1) edge [black, bend right=70,looseness=1.2] node[sloped,rotate=180] {\small{$k-1:\epsilon$}} node[] {\dots} (s1_prime)
        
        (sn_prime) edge [gray,dashed] node[] {}(s_n-2)
        (s_n-1_prime) edge [gray,dashed] node[] {}(dots1)
        (s1_prime) edge [gray,dashed,bend left] node[] {}(term2)
        (s1_prime) edge [gray,dashed,bend right] node[] {}(term2)
        (dots2) edge [gray,dashed] node[] {}(s2_prime)
        (dots2) edge [gray,dashed] node[] {}(s1)
        (s2_prime) edge [gray,dashed] node[] {}(term1)
        ;
    %% -----------------------

    \normalsize

  \end{tikzpicture}
\end{adjustbox} 
\vspace{-0.4cm}
    \caption{The stochastic MDP $H(n, k)$. The original construction of Melekopoglou and Condon is obtained by setting $k = 2$ and $\epsilon = 0$. Edges are labeled ``action: reward''. The introduction of $k - 2$ new actions and related details are presented in Section~\ref{sec:spi_construction}.}
    \label{fig:spi_construction}
\vspace{-0.4cm}
\end{figure*}

%% file: root.bbl
\begin{thebibliography}{10}

\bibitem{Bellman1957}
Richard Bellman.
\newblock {\em Dynamic Programming}.
\newblock Princeton University Press, Princeton, NJ, USA, 1st edition, 1957.

\bibitem{Puterman1994}
Martin~L. Puterman.
\newblock {\em Markov Decision Processes}.
\newblock Wiley, 1994.

\bibitem{Littman+DK:1995}
Michael~L. Littman, Thomas~L. Dean, and Leslie~Pack Kaelbling.
\newblock On the complexity of solving {M}arkov decision problems.
\newblock In {\em Proc. UAI 1995}, pages 394--402. Morgan Kaufmann, 1995.

\bibitem{howard:dp}
R.~A. Howard.
\newblock {\em Dynamic Programming and Markov Processes}.
\newblock MIT Press, Cambridge, MA, 1960.

\bibitem{mansour_singh}
Yishay Mansour and Satinder Singh.
\newblock On the complexity of policy iteration.
\newblock In {\em Proc. UAI 1999}, pages 401--408. Morgan Kaufmann, 1999.

\bibitem{Taraviya2019}
Meet Taraviya and Shivaram Kalyanakrishnan.
\newblock A tighter analysis of randomised policy iteration.
\newblock In {\em Proc. {UAI} 2019}, page ID 174. {AUAI} Press, 2019.

\bibitem{Kalyanakrishnan+MG:2016}
Shivaram Kalyanakrishnan, Neeldhara Misra, and Aditya Gopalan.
\newblock Randomised procedures for initialising and switching actions in
  policy iteration.
\newblock In {\em Proc. AAAI 2016}, pages 3145--3151. AAAI Press, 2016.

\bibitem{Melekopoglou1994}
Mary Melekopoglou and Anne Condon.
\newblock On the complexity of the policy improvement algorithm for {M}arkov
  decision processes.
\newblock {\em INFORMS Journal on Computing}, 6:188--192, 1994.

\bibitem{HZ2010}
Thomas~Dueholm Hansen and Uri Zwick.
\newblock Lower bounds for {H}oward's algorithm for finding minimum mean-cost
  cycles.
\newblock In {\em Algorithms and Computation}, pages 415--426. Springer, 2010.

\bibitem{Fearnley2010}
John Fearnley.
\newblock Exponential lower bounds for policy iteration.
\newblock In {\em Proc. ICALP 2010}, pages 551--562. Springer, 2010.

\bibitem{Hollanders+DJ:2012}
Romain Hollanders, Bal{\'{a}}zs Gerencs{\'{e}}r, and Jean{-}Charles Delvenne.
\newblock The complexity of policy iteration is exponential for discounted
  {M}arkov decision processes.
\newblock In {\em Proc. CDC 2012}, pages 5997--6002. IEEE, 2012.

\bibitem{Bertsekas+Tsitsiklis:1996}
Dimitri~P. Bertsekas and John~N. Tsitsiklis.
\newblock {\em Neuro-Dynamic Programming}.
\newblock Athena Scientific, 1996.

\bibitem{Mahadevan:1996}
Sridhar Mahadevan.
\newblock Average reward reinforcement learning: Foundations, algorithms, and
  empirical results.
\newblock {\em Machine Learning}, 22(1--3):159--195, 1996.

\bibitem{Schurr+Szabo:2005}
Ingo Schurr and Tibor Szab{\'o}.
\newblock Jumping doesn't help in abstract cubes.
\newblock In {\em Integer Programming and Combinatorial Optimization}, pages
  225--235. Springer, 2005.

\end{thebibliography}
